\documentclass{article}

\usepackage{hyperref}

\usepackage[accepted]{icml2023}

\usepackage{algorithm}
\usepackage{algorithmic}
\usepackage{xcolor}

\usepackage{graphicx}
\usepackage{amsmath}
\usepackage{amssymb}
\usepackage{booktabs}

\usepackage{amsmath}
\usepackage{amssymb}
\usepackage{mathtools}
\usepackage{amsthm}

\usepackage{pgfplots}
\usepackage{xfrac}
\usepackage{enumitem}

\usepackage{commath}
\usepackage{amsmath}
\usepackage{pgfplots}
\usepackage{xfrac}
\usepackage{amsfonts}
\usepackage{enumitem}
\usepackage{physics}
\usepackage{todonotes}
\usepackage{adjustbox}
\usepackage{multirow}
\usepackage{makecell}
\usepackage{filecontents}

\usepackage{fancyhdr}

\fancypagestyle{plain}{
  \fancyhf{} 
  \renewcommand{\headrulewidth}{0pt} 
  \renewcommand{\footrulewidth}{0pt} 
  \fancyfoot[C]{\raisebox{-1cm}{\thepage}} 
}
\usepackage[capitalize]{cleveref}
\crefname{section}{Sec.}{Secs.}
\Crefname{section}{Section}{Sections}
\Crefname{table}{Table}{Tables}
\crefname{table}{Tab.}{Tabs.}

\usepgfplotslibrary{fillbetween}
\usepgfplotslibrary{groupplots}

\pgfplotsset{compat=1.12}
\theoremstyle{plain}

\theoremstyle{definition}

\theoremstyle{remark}

\newcommand{\DRIVE}{\texttt{\textsc{DRIVE}}}

\icmltitlerunning{\DRIVE{}: Dual Gradient-Based Rapid Iterative Pruning}

\begin{document}

\twocolumn[
\icmltitle{\DRIVE{}: Dual Gradient-Based Rapid Iterative Pruning}

\icmlsetsymbol{equal}{*}

\begin{icmlauthorlist}
\icmlauthor{Dhananjay Saikumar}{yyy}
\icmlauthor{Blesson Varghese}{yyy}
\end{icmlauthorlist}

\icmlaffiliation{yyy}{\textit{School of Computer Science, University of St Andrews, UK}}
\icmlcorrespondingauthor{Dhananjay Saikumar}{ds304@st-andrews.ac.uk}
\icmlcorrespondingauthor{Blesson Varghese}{bv6@st-andrews.ac.uk}

\vskip 0.3in
]

\printAffiliationsAndNotice{} 

\begin{abstract}
Modern deep neural networks (DNNs) consist of millions of parameters, necessitating high-performance computing during training and inference. Pruning is one solution that significantly reduces the space and time complexities of DNNs. Traditional pruning methods that are applied post-training focus on streamlining inference, but there are recent efforts to leverage sparsity early on by pruning before training. Pruning methods, such as iterative magnitude-based pruning (IMP) achieve up to a 90\% parameter reduction while retaining accuracy comparable to the original model. However, this leads to impractical runtime as it relies on multiple train-prune-reset cycles to identify and eliminate redundant parameters. In contrast, training agnostic early pruning methods, such as SNIP and SynFlow offer fast pruning but fall short of the accuracy achieved by IMP at high sparsities. To bridge this gap, we present Dual Gradient-Based Rapid Iterative Pruning (\DRIVE{}), which leverages dense training for initial epochs to counteract the randomness inherent at the initialization. Subsequently, it employs a unique dual gradient-based metric for parameter ranking. It has been experimentally demonstrated for VGG and ResNet architectures on CIFAR-10/100 and Tiny ImageNet, and ResNet on ImageNet that DRIVE consistently has superior performance over other training-agnostic early pruning methods in accuracy. Notably, \DRIVE{} is 43$\times$ to 869$\times$ faster than IMP for pruning.
\end{abstract}

\section{Introduction}
\label{sec:introduction}
Pruning deep neural networks (DNNs) reduces the number of parameters in the network. The feasibility of pruning without significantly affecting accuracy has been known for more than three decades~\cite{Skeletonization, brain_damage, brain_surgeon, 1993_pruning_survey}. One benefit of pruning is reducing the computational costs during inference~\cite{SparseDNN, PruneFL, li2017pruning, dong_layer_wise_brain, zeng2019mlprune} when combined with the use of appropriate software libraries~\cite{sparse_library} and specialized hardware that leverage sparse matrix operations~\cite{sparse_FPGA, sparse_A100}. Typically, pruning occurs after training~\cite{brain_damage, hanprune, DBLP:conf/iclr/MolchanovTKAK17, NIPS2016_6e7d2da6, Carreira_2018_CVPR, NIPS2016_2823f479} or gradually during training~\cite{You2020Drawing, late_training_gupta, late_train_Gale, dynamic, PruneFL} to produce a sparse DNN that offers low-cost inference. However, the computational costs of training modern production-level DNNs with complex architectures are substantially high~\cite{energy_cost_dnn, state_of_pruning}. Therefore, more recent research has considered whether pruning can be applied early (i.e., before training), since doing so may reduce training costs~\cite{lth,lthscale,Pre_Defined_Sparse, jorge2021progressive}.

\begin{figure}[t]
  \centering
  \includegraphics[width=0.45\textwidth]{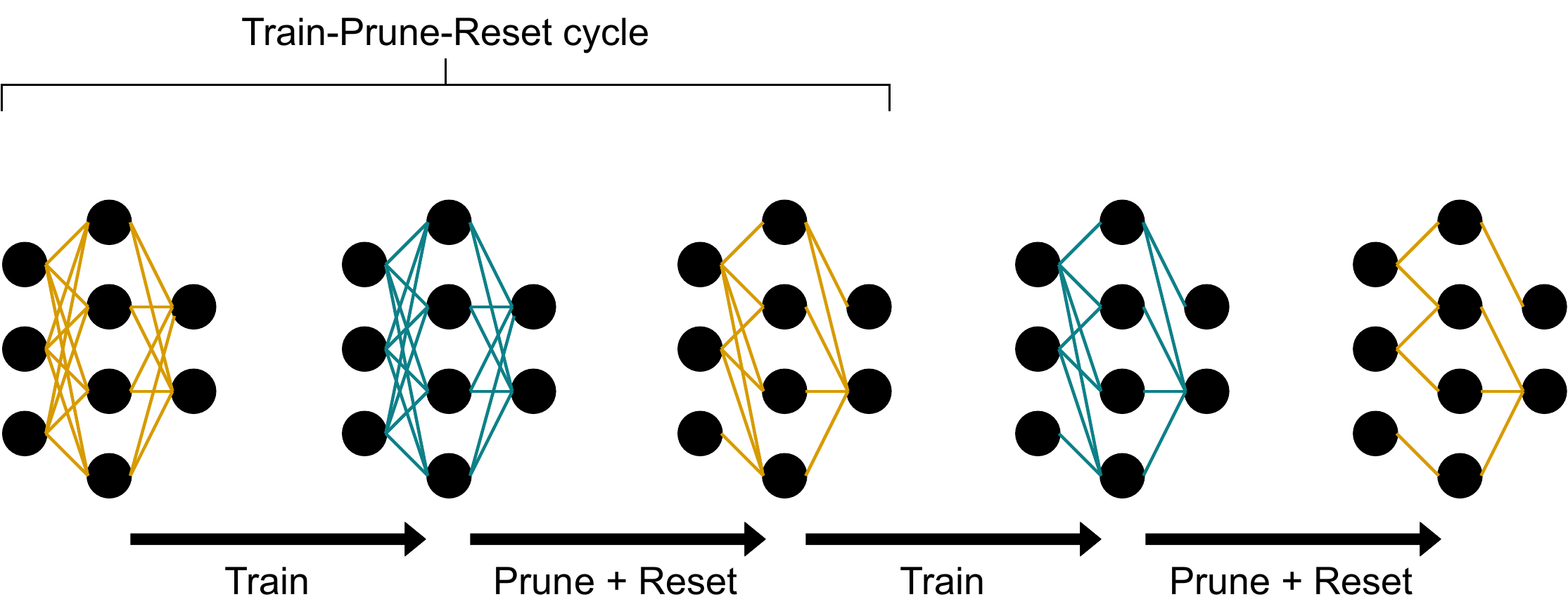}

\caption{Progression of Iterative Magnitude Pruning (IMP) that starts from a dense DNN and has successive train-prune-reset cycles. Neurons are shown as circles, yellow lines indicate weights at random initialization and teal lines show weights post-training.}
\label{fig:lth}
\end{figure}

The catalyst for research into early pruning was the lottery ticket hypothesis (LTH)~\cite{lth, Linear_Mode_Connectivity}. It hypothesizes the existence of certain sub-networks within a DNN, which when trained in isolation match the accuracy of the original dense network. It has been demonstrated that the sub-networks are as sparse as the networks produced by post training pruning methods~\cite{Renda2020Comparing}. 
Thus, high sparsity levels can be maintained throughout training, thereby requiring fewer resources for training and inference.

However, a challenge emerges: While LTH highlights that high-performing sub-networks exist, the method to identify them is underpinned by \textit{iterative magnitude-based pruning} (IMP)~\cite{lth, Linear_Mode_Connectivity}. IMP begins by training the network, pruning its parameters, and then resetting the remaining parameters to their original values, as shown in Figure~\ref{fig:lth}. The train-prune-reset cycle repeats until the desired sparsity is achieved and then a sparse trainable network is returned.
This is computationally intensive and reduces the runtime benefits of subsequently training a sparser model. Therefore, IMP is less suitable for large-scale and on-demand pruning.

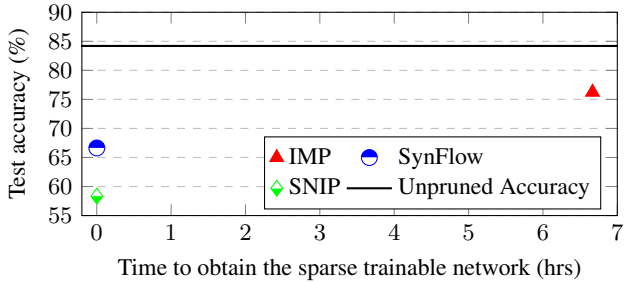
\begin{figure}[t]
\centering
\begin{tikzpicture}[trim axis left,trim axis right,scale=1]
    \begin{axis}[
        width= 8.7 cm,
        height=0.25\textwidth, 
        xlabel={Time to obtain the sparse trainable network (hrs)},
        ylabel={Test accuracy (\%)},
        xmin=-0.2, xmax=7,  
        ymin=55, ymax=90,
        xtick={0,1,2,3,4,5,6,7},
        ytick={55,60,65,70,75,80,85,90},
        legend pos=south east,  
        ymajorgrids=true,
        grid style=dashed,
        legend style={font=\footnotesize, cells={anchor=west}, inner sep=2pt,legend columns=2},  
        tick label style={font=\footnotesize},
        label style={font=\footnotesize},
        legend cell align=left,
        mark options={scale=1.5},  
    ]

    \addplot[
        only marks,
        color=red,
        mark=triangle*,
        ]
        coordinates {
        (24000/3600,76.24)};  
    \addlegendentry{IMP}
        
    \addplot[
    only marks,
    color=blue,
    mark=halfcircle*,
    ]
    coordinates {
    (4.2/3600,66.64)};  
    \addlegendentry{SynFlow}
    
    \addplot[
    only marks,
    color=green,
    mark=halfdiamond*,
    ]
    coordinates {
    (3.13/3600,58.35)};  
    \addlegendentry{SNIP}

    \addplot [mark=none,black,thick] coordinates {
    (-0.5,84.2) (8,84.2)};
    \addlegendentry{Unpruned Accuracy}

    \end{axis}
\end{tikzpicture}

\caption{Comparing test accuracy of sparse networks derived using early pruning methods for AlexNet with 99.3\% parameters removed and trained on the CIFAR-10 dataset. 
}
\label{fig:existing_techniques}
\end{figure}

To surmount the above challenge, pruning research has recently considered  efficiently identifying sub-networks at initialization without any training. Notably, SynFlow, using an iterative method, prunes based on the smallest `synaptic saliency'~\cite{SynFlow}, while SNIP removes parameters that have a minimal impact on the loss when removed~\cite{snip}. However, while pruning-at-initialization methods are faster, they produce sub-networks with lower accuracy than IMP~\cite{missing_mask_pruning, Linear_Mode_Connectivity, state_of_pruning} at extreme sparsities. This is highlighted in Figure~\ref{fig:existing_techniques}, where AlexNet~\cite{alexnet} is pruned to a sparsity of 99.3\% and trained on CIFAR-10~\cite{CIFAR10}. While SNIP and SynFlow require less than a minute for pruning, the resulting sub-networks lag behind IMP in accuracy by 18\% and 10\%, respectively. IMP, requires a longer pruning time but yields sub-networks with higher accuracy. This highlights the trade-off between speed and performance that will potentially be accentuated as dataset and network complexities increase. Though IMP can derive high-performing sub-networks even at extreme sparsities, methods like SNIP and SynFlow do not. This highlights the potential for improving early pruning methods, which our work taps into.

In this paper, we introduce the \textbf{\textit{Dual Gradient-Based Rapid Iterative Pruning (\DRIVE{})}}, designed to bridge the performance gap between exhaustive and initialization-based pruning methods. \DRIVE{} starts by training the unpruned model for a few epochs before initiating the pruning process, thus accessing only partial information compared to methods, such as IMP that require exhaustive training to determine parameter importance. To compensate, \DRIVE{} utilizes a distinctive dual gradient metric to rank parameters before iteratively pruning them. This metric incorporates three salient terms: the \textit{parameter magnitude}, the \textit{connection sensitivity}, and the \textit{convergence sensitivity}. Beyond evaluating and pruning based on the current importance  of the parameters (magnitude that is akin to IMP), the \DRIVE{} metric avoids premature pruning of currently less relevant parameters that may be required in the future. The outcome is a sparse, trainable network that achieves the efficiency of initialization-based methods while bridging the performance gap of exhaustive pruning methods, namely IMP. Across diverse architectures such as AlexNet, VGG-16, and ResNet-18, and datasets, namely CIFAR-10, CIFAR-100,  Tiny ImageNet, and ResNet-18 on ImageNet, \DRIVE{} consistently surpasses the performance of training-agnostic pruning methods, namely SNIP and SynFlow. Furthermore, \DRIVE{} boasts an enhancement in computational efficiency of atleast, 43$\times$ and, at best, 869$\times$ faster than IMP.

Our main contributions are:
\begin{itemize}[noitemsep,nosep]
    \item \textbf{Development of \DRIVE{}}: A novel pruning method that bridges the advantages of initialization-based and exhaustive training-dependent pruning methods.
    
    \item \textbf{Efficient early pruning with \DRIVE{}}: Unlike exhaustive methods, such as IMP, \DRIVE{} streamlines pruning as it requires training only for a few epochs.
    
    \item \textbf{Novel dual gradient metric}: We present a novel metric considering parameter magnitude, connection sensitivity, and proximity to convergence, to ensure optimal pruning decisions.
\end{itemize}

The rest of this paper is organised as follows: Section~\ref{sec:relatedwork} reviews the state-of-the-art. Section~\ref{sec:DGP} presents the \DRIVE{} method. Section~\ref{sec:studies} presents the results of experiments comparing the performance of \DRIVE{} to the unpruned network and existing pruning methods such as IMP, SNIP, and SynFlow. We conclude the paper in Section~\ref{sec:discussion}.

\section{Background \& Related Work}
\label{sec:relatedwork}
This section presents the mathematical formulation of pruning, the existing pruning methods, and the limitations of pruning at initialization.

\textbf{Formulating Pruning}: 
Given a DNN $f(\theta;x)$ with parameters $\theta$ and input $x$, pruning produces a sub-network with a set of parameters $\theta \odot m$, where \(m \in \{0,1\}^{d}\) is a binary mask, and \(\odot\) is the element-wise product. The degree of sparsity, \(\kappa \in [0,1]\), indicates the fraction of parameters removed from the original network to produce the sub-network. All pruning methods assign scores \(z \in \mathbb{R}^{d}\) to the parameters \(\theta\) to rank parameters to identify which ones can be discarded. 

Pruning is carried out in the following two ways: (1) \textit{One-shot pruning} by assigning scores and pruning parameters in a single step, resulting in a final sparsity of~$\kappa$. (2) \textit{Iterative pruning} by repeatedly assigning scores and pruning parameters over a series of \( N \) iterations. At each iteration \( n \), the network is pruned to achieve a sparsity level \( \kappa_n \), where \( \kappa_n \) is a pre-defined schedule that monotonically increases from a low value to the final desired sparsity \( \kappa \).

\textbf{Early Pruning Methods}:
We now describe the IMP method and two early pruning methods from the literature, namely SNIP and SynFlow.

\textit{IMP}: For a desired sparsity value \( \kappa \), and an initialized DNN \( f(\theta^{0};x) \), IMP starts by training the DNN for \( E \) epochs, transforming it from \( f(\theta^{0};x) \) to \( f(\theta^{E};x) \). After training, parameters are assigned scores, represented by \( z = |\theta^{E}| \). \( \rho\% \) of the parameters are pruned, specifically those with the lowest scores: 
\begin{equation}\label{eq:prune_iter}
\rho = \kappa^{1/N}
\end{equation}
Once pruned, the surviving parameters are reset to their initial values. This entire process is repeated for \( N \) cycles to achieve the target sparsity \( \kappa \)~\cite{lth, lthscale}.

\textit{SNIP}: For a desired sparsity value of \( \kappa \), and an initialized DNN \( f(\theta^{0};x) \), SNIP evaluates the connection sensitivity of each parameter within the DNN relative to a training loss $L$. A mini-batch, $B$ is sampled from the training dataset. The primary computation in SNIP is to determine the gradient of the loss $L$ with respect to each connection mask $m_{j}$, which captures the effect on the loss when the parameter is removed from the network. 
\begin{equation}\label{eq:unnorm_connection}
g_j(\theta^{0}; B) = \frac{\partial L(\theta^{0} \odot m; B)}{\partial m_j}
\end{equation}
Using these gradients, the score \( z_j \) for each parameter is computed as the normalized magnitude of the gradients:
\begin{equation}\label{eq:norm_connection}
z_j = \frac{|g_j(\theta^{0}; B)|}{\sum_{k=1}^{m} |g_k(\theta^{0}; B)|}
\end{equation}

The bottom \( \kappa \% \) of parameters, in terms of their scores, are pruned in a one-shot manner~\cite{snip}. This ensures that only parameters with the highest influence on the loss, whether positive or negative, are preserved~\cite{missing_mask_pruning}.

\textit{SynFlow}: For a desired sparsity value of \( \kappa \), and an initialized DNN \( f(\theta^{0};x) \), SynFlow begins by converting the parameters \( \theta^{0} \) to their absolute values \( |\theta^{0}| \). The DNN is then evaluated using an input matrix where every element is set to one (denoted as \( \mathbb{I} \)). The resulting scalar value \( A \) is:
\begin{equation}\label{eq:sum_logits}
    A = \sum f(|\theta^{0}|;\mathbb{I})
\end{equation}
Following this, scores \(z\) are assigned as:
\begin{equation}\label{eq:synflow_score}
z = \frac{\partial A}{\partial \theta^{0}} \odot \theta^{0}
\end{equation}

Using Equation~\ref{eq:prune_iter}, \( \rho\% \) of the parameters, specifically those with the lowest scores, are pruned from the network. Issuing scores and pruning are repeated over $N$ iterations~\cite{SynFlow,missing_mask_pruning}. 
SynFlow was designed to prevent `layer collapse', which occurs when all parameters in a single layer are pruned, thus making the network untrainable~\cite{SynFlow,layer_collapse2,layer_collapse3}. To address this, SynFlow ensures the layer-wise conservation of the scores \(z\) in a data-agnostic and iterative fashion.

\textbf{Limitations of Pruning at Initialization}:
Parameters in a DNN are assigned random values at initialization, providing no valuable insight at this stage as to which parameters are of significance to a specific DNN-dataset combination. As a result, IMP~\cite{lth} and magnitude-based pruning after training~\cite{hanprune} leverage training to determine essential parameters, but they are computationally intensive. In contrast, pruning-at-initialization methods, such as SNIP and SynFlow, offer faster alternatives at the expense of a performance gap considered in Section~\ref{sec:introduction}. The reasons for the performance gap are discussed next.

\begin{figure}[t]
\centering
\begin{tikzpicture}[trim axis left,trim axis right,scale=1]
    \begin{axis}[
        width= 8.7 cm,
        height=0.24\textwidth, 
        xlabel={Network sparsity ($\kappa$\%)},
        ylabel={$\Delta$ Accuracy (\%)},
        ylabel shift=-5pt, 
        xmin=95, xmax=99.5,
        ymin=-70, ymax=5,
        ymajorgrids=true,
        grid style=dashed,
        legend pos=south west,
        legend style={font=\footnotesize, cells={anchor=west}, inner sep=2pt,legend columns=2},
        tick label style={font=\footnotesize},
        label style={font=\footnotesize},
        legend cell align=left,
        mark options={scale=1.5}
    ]
    
    \addplot[
        color=green,mark=halfdiamond*
        ]
        coordinates {
(90,64.44-64.05)
(95,62.9-63.81)
(97,59.7-62.56)
(98,58.38-61.94)
(99,48-62.91)
(99.3,33.97-61.08)
(99.5,1-60.72)

};
\addlegendentry{SNIP}

\addplot[
        color=blue,mark=halfcircle*
        ]
        coordinates {
(90,64.44-63.29)
(95,62.9-61.02)
(97,59.04-62.56)
(98,56.59-61.94)
(99,53.14-62.91)
(99.3,51.27-61.08)
(99.5,47.82-60.72)

};
\addlegendentry{SynFlow}

\addplot [black,dashed] coordinates {
(90,0) (99.5,0)
};
\addlegendentry{IMP (Reference)}

\end{axis}
\end{tikzpicture}

\caption{Change in test accuracy of trained sparse networks produced by SNIP and SynFlow relative to IMP, plotted against the network sparsity for ResNet-18 on the CIFAR-100 dataset.}
\label{fig:relative_acc}
\end{figure}
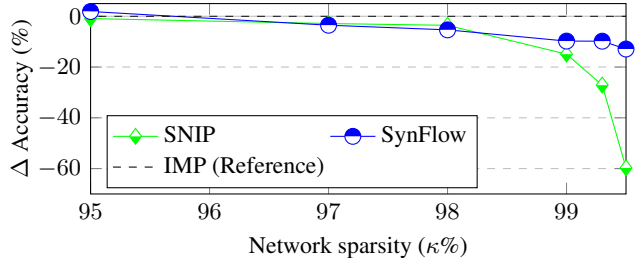

SNIP ranks parameters by evaluating their impact on the loss \(L\) when removed, and is captured by the transition of a binary mask from \(m_{j} = 1\) to \(m_{j} = 0\). To replicate this transition, SNIP computes the gradient \(\frac{\partial L}{\partial m_{j}}\) shown in Equation~\ref{eq:unnorm_connection}, and scores parameters based on the \textbf{normalized} magnitude of these derivatives. Then \( \kappa \% \) of the parameters with the lowest scores are subsequently pruned in a one-shot manner~\cite{snip}. A critical limitation of SNIP is its scoring mechanism, which considers only the \textbf{absolute} impact of parameters on the loss $L$, not distinguishing between positive or negative impact. Due to randomized initialization, several non-essential parameters may amplify the loss $L$, leading them to achieve high scores and remain unpruned. When compared to IMP, this method adversely affects accuracy, specifically in high-sparsity settings, as shown in Figure~\ref{fig:relative_acc}. As a result, sub-networks produced by SNIP are consistently inferior to those from IMP. Additionally, SNIP is also vulnerable to layer collapse at higher sparsity levels~\cite{SynFlow}, as evident from the sharp drop in accuracy shown in Figure~\ref{fig:relative_acc}.

SynFlow leverages a data-agnostic scoring mechanism that preserves scores for parameters in a layer-centric manner. This preserves inter-layer connections, thereby acting as a safeguard against layer collapse. However, despite its demonstrated superiority over SNIP in mitigating layer collapse, a notable limitation arises from its data-agnostic scoring approach. Specifically, SynFlow cannot determine whether parameters are essential to a specific DNN-dataset combination. To illustrate, given a DNN and two datasets that are dimensionally identical but have drastically different features, SynFlow would yield the same sub-networks for both. This leads to suboptimal performance, particularly at higher sparsity levels. While the performance of SynFlow is better than SNIP, it is often lower than IMP, as evident in Figure~\ref{fig:relative_acc}. 

\textbf{Gradual Pruning Methods}: The above methods generate trainable sparse networks in which the binary mask remains static during training. However, there is another class of methods, referred to as gradual pruning~\cite{late_training_gupta, prunetrain, You2020Drawing, dynamic, ClickTrain}, which incrementally prune the network during training. While this method is more cost-effective than pruning after completing training, it still requires substantial training to identify the optimal sparse sub-network. Our research is centred on early identification and training of sparse sub-networks with high sparsity, a method that is more cost-effective and distinct from gradual pruning.

\section{DRIVE Design}
\label{sec:DGP}

In this section, we introduce the Dual Gradient-Based Rapid Iterative Pruning (\DRIVE{}) method, an early two-stage pruning technique. Initially, the model undergoes brief training to address limitations of early pruning, as detailed in Section~\ref{sec:relatedwork}. Then, we iteratively prune parameters to achieve a sparse, trainable model. We will outline the parameter ranking criteria and provide an overview of \DRIVE{} in Algorithm~\ref{alg:dgp}.

\subsection{Pruning Metric}
The choice of a pruning metric is crucial as it determines the hierarchical ranking of parameters in the network. The IMP method uses the L1 norm of the parameter. SNIP evaluates the impact on the loss when a parameter is removed, as presented in Equation~\ref{eq:norm_connection}. SynFlow employs data-agnostic synaptic saliency  as shown in Equation~\ref{eq:synflow_score}. The pruning metric of \DRIVE{} combines the L1 norm of the parameter with two distinct gradient-based terms as shown below: 
\begin{multline}\label{eq:DGP_metric}
S_{j}(\theta;B) = \\ \theta_{j} \odot \frac{ \partial L(f(\theta \odot m;x),y)}{\partial m_{j}} \odot \frac{\partial L (f(\theta \odot m;x),y)}{\partial \theta_{j}}
\end{multline}

Within this context, $S_{j}$ denotes the \DRIVE{} metric corresponding to the parameter $\theta_{j}$, with $B$ symbolizing a training batch. Parameters are ranked based on their respective scores, $z_{j}$, which are determined by the normalized magnitude of the \DRIVE{} metric:
\begin{equation}\label{eq:DGP_norm}
z_j = \frac{\abs{S_{j}(\theta;B)}}{\sum_{k=1}^{m} \abs{S_{k}(\theta;B)}}
\end{equation}

The rationale behind the inclusion of each term in the \DRIVE{} metric is considered next.

\begin{figure}[]
  \centering
  \includegraphics[width=0.45\textwidth]{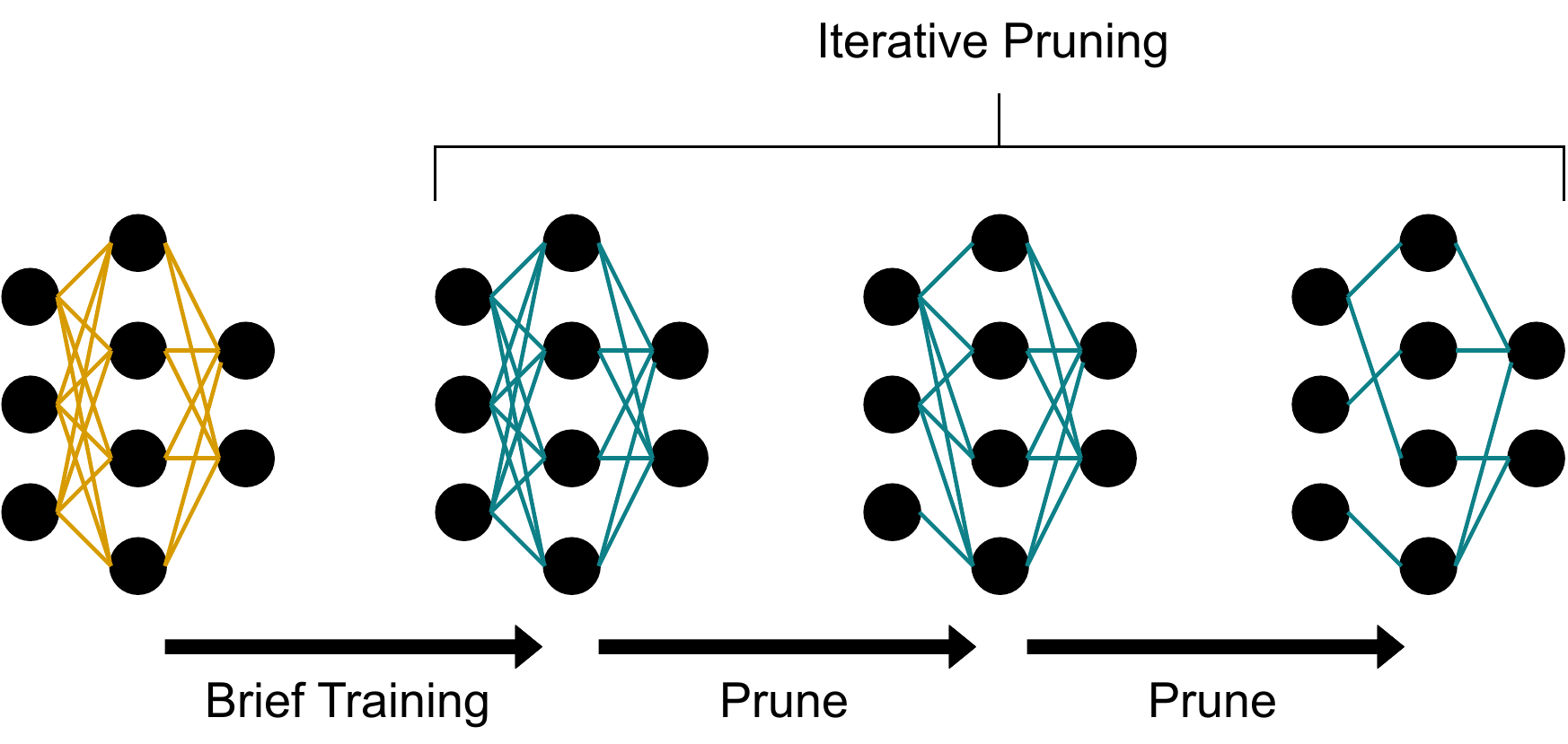}
  \caption{\DRIVE{} initially trains an unpruned network and then iteratively prunes the network. Neurons are circles. Yellow lines are initial weights and teal lines are weights after brief training.}
  \label{fig:dgp}
\end{figure}

\textbf{L1 Norm}:
At initialization, parameter values are drawn randomly, meaning that the L1 norm (magnitude) of these parameters initially offers no insight into their relevance to the network. To address this,
\DRIVE{} begins by training the unpruned model for a few epochs. This step allows essential parameters to acquire larger magnitudes, indicating their significance, while less important parameters tend to acquire smaller values. Following the brief training period, the L1 norm of these parameters contributes to how the parameters are ranked before pruning. 

While IMP subjects the model to a more intensive training period to classify and prioritize parameters using the L1 norm before pruning, \DRIVE{} operates with partial information relative to IMP due to undergoing brief training. To compensate for this shorter training period, \DRIVE{} integrates two supplementary gradient-based metrics along with the L1 norm, which are described next.

\textbf{Connection sensitivity}:
Let $\theta_{j}$ represent the \(j^{th}\) parameter in the network \(f(\theta;x)\). In a given DNN-dataset combination, besides using the L1 Norm, the impact of a parameter \(\theta_{j}\) can be determined by assessing the change in loss when removed from the network. The change in loss, referred to as \textit{connection sensitivity}, captures how sensitive the performance of the network is to the presence or absence of the parameter \(\theta_{j}\)~\cite{snip, GraSP}. 
In mathematical terms, to eliminate the parameter $\theta_{j}$ from the network, its value can be set to zero, or a mask ($\mathbb{I} - e_{j}$) can be applied. Here, $e_{j}$ is the indicator vector for the parameter $\theta_{j}$, being zero-filled except at the \(j^{th}\) index, where it takes the value one. The matrix $\mathbb{I}$, functioning as a mask, has entirely values of one. The change in loss when removing $\theta_{j}$ is calculated in two forward passes:
\begin{multline}\label{eq:6}
\Delta L_{j}(f(\theta;x),y) = \\ L(f(\theta \odot \mathbb{I};x),y) - L(f(\theta \odot (\mathbb{I} - e_{j});x),y)
\end{multline}

However, given that modern DNNs have millions of parameters, computing the connection sensitivity for each individual parameter is computationally intensive since millions of forward passes would be required. A more efficient approach is to approximate the effect on the loss when a parameter $\theta_{j}$ is excluded. This approximation, as illustrated in SNIP~\cite{snip}, is shown as:
\begin{multline}\label{eq:7}
\Delta L_{j}(f(\theta;x),y) \approx \\ \lim_{\delta\to 0} \frac{L(f(\theta \odot \mathbb{I};x),y) - L(f(\theta \odot (\mathbb{I} - \delta e_{j});x),y)}{\delta} \\ = \frac{\partial L(f(\theta \odot m;x),y)}{\partial m_{j}}\Bigg|_{m = \mathbb{I}}
\end{multline}

The connection sensitivity can be approximated for all parameters in the network efficiency via a single forward-backward pass. Notably, the second term in the \DRIVE{} metric, as shown in Equation~\ref{eq:DGP_metric} describes the connection sensitivity. A larger connection sensitivity value for a parameter implies a significant alteration in loss upon its removal, emphasizing its essential role in the given DNN-dataset combination.

However, when connection sensitivity is calculated is crucial. 
In SNIP, it is calculated at initialization but in \DRIVE{} it is after a period of preliminary training. The model undergoes initial training epochs before any pruning in \DRIVE{}, which guides the parameters to minimize the loss, allowing essential parameters to obtain larger magnitudes. 
If such parameters are removed, the loss changes would be significant. Hence, they are assigned higher sensitivity scores. Conversely, the removal of non-essential parameters, which have been guided to smaller values through the preliminary training, has a limited effect on the loss. As a result, these non-essential parameters are assigned lower sensitivity scores. In contrast, as previously discussed, SNIP evaluates parameters without distinguishing between their positive or negative impact on loss. This is challenging at initialization since many non-essential parameters may have a large impact on the loss. Therefore, SNIP may assign them as crucial. Without the guidance of initial training, SNIP is subject to the inherent randomness of initialization, leading to the potential preservation of non-essential parameters. 

\textbf{Convergence Sensitivity}: The \DRIVE{} approach to pruning is distinct from that of IMP, which relies on train-prune-reset cycles, allowing the network to experience extensive training before pruning takes place. In comparison, \DRIVE{} prunes after a few initial training epochs, thereby considering the emerging state of the network rather than its mature state following in-depth training. The L1 Norm and Connection Sensitivity can only determine the current importance of parameters in this context. However, we note that parameters exist within the network that are not deemed currently impactful but may be important as training progresses. This observation leads to the introduction of the third term in the \DRIVE{} metric.

DNN training updates parameter that are guided by the gradient of the loss with respect to each parameter:
\begin{equation}\label{eq:8}
\Delta \theta = - \eta \frac{\partial L}{\partial \theta}
\end{equation}
Here $\eta$ represents the learning rate.
As conventional DNN training progresses, the sequence of expected gradient ($\frac{\partial L}{\partial \theta}$) norm approaches zero~\cite{Optimization_Methods}. Therefore, the magnitude of the gradient in Equation~\ref{eq:8} serves as a proxy for the proximity of the parameter to convergence. We refer to this gradient as \textit{convergence sensitivity}. A large convergence sensitivity value suggests that a parameter is far from convergence, and requiring further training to determine its significance. Parameters with large convergence sensitivity scores are assigned a higher priority to ensure that they are not pruned prematurely.

\begin{algorithm}[h]
   \caption{Dual Gradient-Based Rapid Iterative Pruning}
   \label{alg:dgp}
\begin{algorithmic}[1]
   \STATE {\bfseries Input:} Training dataset $\mathcal{D}$, Targeted sparsity $\kappa$, Pruning cycles $N$, Training epochs $E$, Initial network $f(\theta^{0};x)$
   \STATE {\bfseries Output:} Sparse trainable network $f(\theta^{E} \odot m;x)$
   \vspace{0.5em}
   \STATE \textbf{Early Training Phase:}
   \FOR{$j=1$ {\bfseries to} $E$}
       \STATE Update parameters: $f(\theta^{j+1};x) \leftarrow f(\theta^{j};x)$
   \ENDFOR
   
   \vspace{0.5em}
   \STATE Apply mask ($m = \mathbb{I}$) to the network: $f(\theta^{E} \odot m;x) \leftarrow f(\theta^{E};x)$
   \vspace{0.5em}
   \STATE Calculate pruning fraction: $\rho \leftarrow \kappa^{1/N}$
   \vspace{0.5em}
   \STATE \textbf{Iterative Pruning Stage:}
   \FOR{$j=1$ {\bfseries to} $N$}
       \vspace{0.5em}
       \STATE Sample a training batch $\textbf{B}$ from $\mathcal{D}$
       \vspace{0.5em}
       \STATE Calculate connection sensitivity $\leftarrow$ Eq \ref{eq:7} 
       \vspace{0.5em}
       \STATE Calculate convergence sensitivity $\leftarrow$ Eq \ref{eq:8}
       \vspace{0.5em}
       \STATE Compute \DRIVE{} metric: $S_{j}(\theta;B) = \abs{\theta_{j} \odot \frac{\partial L(f(x;\theta \odot m),y)}{\partial m_{j}} \odot \frac{\partial L(f(x;\theta \odot m),y)}{\partial \theta_{j}}}$
       \vspace{0.5em}
       \STATE Assign scores: $z_j = \frac{S_{j}(\theta;B)}{\sum_{k=1}^{m} S_{k}(\theta;B)}$
       \vspace{0.5em}
       \STATE Shortlist \(\rho\%\) parameters ($\tau$) by lowest score \(z\): 
       $\tau \leftarrow z$
       \vspace{0.5em}
       \STATE Update mask: Set mask elements of $\tau$ to $0$
       \vspace{0.5em}
   \ENDFOR
   \vspace{0.5em}
   \STATE {\bfseries return} $f(\theta^{E} \odot m;x)$
\end{algorithmic}
\end{algorithm}

\section{Experiments}
\label{sec:studies}
In this section, we detail the experimental setup and present the results obtained from our study. 
\subsection{Setup}
We empirically evaluate the performance of \DRIVE{} against the computationally intensive iterative magnitude-based pruning (IMP) and two established early pruning methods, namely SNIP and SynFlow. Our tests involve three DNN architectures: AlexNet~\cite{alexnet} trained on the CIFAR-10 dataset~\cite{CIFAR10}, and VGG-16~\cite{VGG} and ResNet-18~\cite{RESNET}, both trained on CIFAR-10 and CIFAR-100~\cite{CIFAR10}. Additionally, ResNet-18 is evaluated on the ImageNet dataset~\cite{ImageNet}. All measurements were averaged over three runs on an NVIDIA RTX A6000 GPU.


\textbf{Pruning Methods}: 
The pruning methods compared in this study generate trainable sparse networks using static masks that have constant values and remain unchanged during training. This approach contrasts with gradual pruning methods, which incrementally prune the network throughout the training process. Our research is focused on the early identification and training of highly sparse sub-networks, which is more cost-efficient and distinct from gradual pruning techniques. Consequently, \DRIVE{} is compared only with baselines that are early pruning methods, namely IMP, SNIP and SynFlow.

For AlexNet on CIFAR-10, as well as for ResNet-18 and VGG-16 on CIFAR-10/100, Tiny ImageNet, and ResNet-18 on ImageNet, the IMP method involves 50 train-prune-reset cycles, with each cycle consisting of 20 training epochs. SynFlow, maintaining consistency across all architectures and datasets, utilizes 100 pruning iterations. SNIP, by its design, applies a one-shot pruning method to all architectures across all datasets. For the DRIVE algorithm, AlexNet on CIFAR-10, ResNet-18, and VGG-16 on CIFAR-10/100 were pre-trained for 5 epochs followed by 100 pruning iterations. For ResNet-18 and VGG-16 on Tiny ImageNet and ResNet-18 on ImageNet, the procedure initiates with 1 pre-training epoch, followed by 100 pruning iterations. To ensure equal total training time across methods, networks pruned with DRIVE undergo training for five fewer epochs on CIFAR-10/100 and one fewer epoch on Tiny ImageNet and ImageNet.

\textbf{Training Scheme}: 
Across all architectures and datasets, both the unpruned (reference) network and the sparse trainable networks generated by the pruning techniques were subjected to the following training schemes. For AlexNet on CIFAR-10, training lasted 50 epochs, using a batch size of 256, an Adam optimizer, and a learning rate of 0.0001. Transitioning to the VGG-16 and ResNet-18 models on CIFAR-10, they were trained for 150 and 200 epochs, respectively. Both employed a batch size of 128 and the SGD optimizer with a learning rate of 0.01. Specifically for VGG-16, its learning rate was reduced by a factor of 5 every 30 epochs, while for ResNet-18, a Cosine Annealing approach smoothly decayed the rate across 200 epochs.

VGG-16 and ResNet-18 models on CIFAR-100 required 200 epochs of training using a batch size of 128 and an initial learning rate of 0.01. Notably, the learning rate of VGG-16 was dropped by a factor of 5 at epochs 60, 120, and 160. Conversely, the learning rate for ResNet-18 was reduced by a factor of 10 at epoch 80 and further reduced by a factor of 100 at epoch 120.

On the Tiny ImageNet dataset, both VGG-16 and ResNet-18 underwent 100 epochs of training, utilizing a batch size of 128 and an initial learning rate of 0.01 with the SGD optimizer. The learning rates were dropped by a factor of 10 at epochs 30, 60, and 80.

For ResNet-18 on ImageNet, for consistency in benchmarking across the same architecture using CIFAR datasets, ImageNet data samples were downsampled from $224\times224$ to $32\times32$. Training for ResNet-18 on ImageNet lasted 100 epochs, with a batch size of 256 and an initial learning rate of 0.1, employing the SGD optimizer. The learning rates were reduced by a factor of 10 at epochs 30, 60, and 80.

\subsection{Results}

\begin{table*}[t]
\caption{Test accuracies of AlexNet, VGG-16, and ResNet-18 on different datasets. Here the underlined values indicate an untrainable network resulting from layer collapse. Bold values indicate the highest accuracy for the respective sparsity level and model among the early pruning techniques: SNIP, SynFlow, and \DRIVE{}.}
\begin{center}
\scalebox{0.9}{
\fontsize{8.5pt}{11pt}\selectfont 
\begin{tabular}{l|l|l|cccccc}
\hline
\textbf{Dataset} & \textbf{Model} & \textbf{Sparsity} & \textbf{90\%} & \textbf{95\%} & \textbf{98\%} & \textbf{99.3\%} & \textbf{Pruning time (seconds)} \\
\hline
\multirow{15.1}{*}{CIFAR-10} 
& \multirow{5}{*}{\textbf{AlexNet}} & Unpruned & \multicolumn{4}{c}{84.20} & & \\
& & IMP & 85.76 & 83.58 & 80.57 & 76.24 & 24$\times 10^3$ \\

& & SNIP & 83.09 & 79.95 & 72.95 & 58.35 & 3 \\
& & SynFlow & 86.37 & 84.02 & 78.61 & 66.64 & 4 \\
& & \DRIVE{} & \textbf{86.97} & \textbf{87.21} & \textbf{85.37} & \textbf{74.49} & 220 \\
\cline{3-8}
& \multirow{5}{*}{\textbf{VGG-16}} & Unpruned & \multicolumn{4}{c}{92.40} & & \\
& & IMP & 91.61 & 90.20 & 90.04 & 88.35 & 12$\times 10^3$ \\

& & SNIP & 92.56 & 92.01 & 91.31 & \underline{10.00} & 2 \\
& & SynFlow & 91.32 & 90.44 & 89.79 & 87.63 & 3 \\
& & \DRIVE{} & \textbf{92.80} & \textbf{92.68} & \textbf{91.36} & \textbf{89.77} & 260 \\
\cline{3-8}
& \multirow{5}{*}{\textbf{ResNet-18}} & Unpruned & \multicolumn{4}{c}{93.70} & & \\
& & IMP & 91.79 & 90.81 & 90.71 & 88.66 & 21$\times 10^3$ \\

& & SNIP & 93.56 & 92.03 & 89.66 & 84.56 & 2 \\
& & SynFlow & 91.05 & 90.23 & 88.74 & 82.62 & 3 \\
& & \DRIVE{} & \textbf{93.59} & \textbf{92.14} & \textbf{89.86} & \textbf{85.90} & 270 \\
\cline{3-8}
\hline
\multirow{11}{*}[1.4ex]{CIFAR-100} 
& \multirow{5}{*}{\textbf{VGG-16}} & Unpruned & \multicolumn{4}{c}{65.50} & & \\
& & IMP & 65.80 & 65.50 & 63.60 & 61.20 & 12.6$\times 10^3$ \\

& & SNIP & \textbf{65.11} & 64.06 & 56.78 & \underline{1.00} & 2 \\
& & SynFlow & 63.72 & 63.02 & 61.45 & 54.24 & 3 \\
& & \DRIVE{} & 65.05 & \textbf{64.36} & \textbf{61.58} & \textbf{59.70} & 290 \\
\cline{3-8}
& \multirow{5}{*}{\textbf{ResNet-18}} & Unpruned & \multicolumn{4}{c}{66.50} & & \\
& & IMP & 64.44 & 63.81 & 61.94 & 61.08 & 22\(\times 10^3\)\\

& & SNIP & 64.05 & 62.90 & 58.38 & 33.97 & 2\\
& & SynFlow & 63.29 & 61.02 & 56.59 & 51.27 & 3 \\
& & \DRIVE{} & \textbf{64.34} & \textbf{63.01} & \textbf{60.30} & \textbf{57.44} & 320 \\
\cline{3-8}
\hline
\multirow{11}{*}[1.4ex]{Tiny ImageNet} 
& \multirow{5}{*}{\textbf{VGG-16}} & Unpruned & \multicolumn{4}{c}{48.74} & & \\
& & IMP & 47.85 & 46.79 & 44.26 & 36.26 & 63\(\times 10^3\) \\

& & SNIP & 44.45 & 39.51 & 33.21 & 16.85 & 2 \\
& & SynFlow & 47.23 & 43.03 & 40.46 & 30.86 & 4 \\
& & \DRIVE{} & \textbf{48.29} & \textbf{45.60} & \textbf{42.25} & \textbf{37.97} & 160 \\
\cline{3-8}
& \multirow{5}{*}{\textbf{ResNet-18}} & Unpruned & \multicolumn{4}{c}{49.41} & & \\
& & IMP & 46.01 & 44.54 & 42.35 & 40.70 & 119\(\times 10^3\) \\

& & SNIP & 44.45 & 39.51 & 33.21 & 16.85 & 2 \\
& & SynFlow & 44.04 & 39.83 & 36.77 & 30.64 & 3 \\
& & \DRIVE{}  & \textbf{45.32} & \textbf{43.89} & \textbf{42.79} & \textbf{33.01} & 190\\
\cline{3-8}
\hline
\end{tabular}}
\end{center}
\label{table:1}
\end{table*}

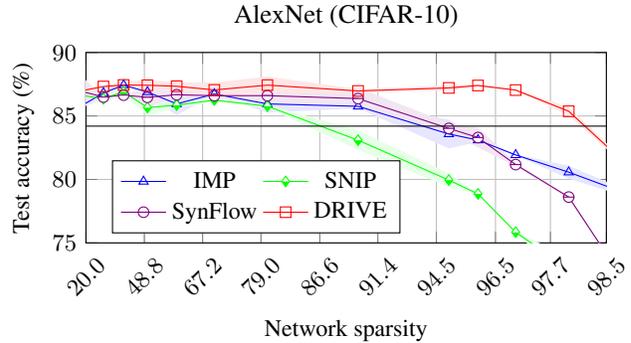
\begin{figure}
\centering
\begin{tikzpicture}
\begin{axis}[
    title= AlexNet (CIFAR-10),
    legend pos=south west,
    legend style={at={(0.05,0.05)}, anchor=south west, legend columns=2,font=\footnotesize},
    x coord trafo/.code={\pgfmathparse{100-ln(100-#1)}},
    x coord inv trafo/.code={\pgfmathparse{100-exp(100-#1)}},
    xmin=20, xmax=98.5,
    ymin=75, ymax=90,
    xmajorgrids=true, ymajorgrids=true,
    xlabel={Network sparsity},
    ylabel={Test accuracy (\%)},
    tick label style={font=\footnotesize},
    label style={font=\footnotesize},
    width = 8.5 cm,
    height=0.24\textwidth, 
    xtick={20.0, 48.8, 67.2, 79.0, 86.6, 91.4, 94.5, 96.5, 97.7,98.5},
    x tick label style={font=\small, rotate=45, /pgf/number format/.cd, fixed, fixed zerofill, precision=1}
]

\addplot [color=blue,mark=triangle] table[x=x,y=y] {ALEXCIFAR10IMP.dat};
\addlegendentry{IMP}

\addplot [color=green,mark=halfdiamond*] table[x=x,y=y] {ALEXCIFAR10SNIP.dat};
\addlegendentry{SNIP}

\addplot [color=violet,mark=halfcircle] table[x=x,y=y] {ALEXCIFAR10SYNFLOW.dat};
\addlegendentry{SynFlow}

\addplot [color=red,mark=square] table[x=x,y=y] {ALEXCIFAR10DGP.dat};
\addlegendentry{DRIVE}

\addplot [mark=none,black] coordinates {(20,84.2) (99.75,84.2)};

\addplot [name path=upper,draw=none] table[x=x,y expr=\thisrow{y}+\thisrow{err}] {ALEXCIFAR10IMP.dat};
\addplot [name path=lower,draw=none] table[x=x,y expr=\thisrow{y}-\thisrow{err}] {ALEXCIFAR10IMP.dat};
\addplot [fill=blue!10] fill between[of=upper and lower];

\addplot [name path=upper,draw=none] table[x=x,y expr=\thisrow{y}+\thisrow{err}] {ALEXCIFAR10SNIP.dat};
\addplot [name path=lower,draw=none] table[x=x,y expr=\thisrow{y}-\thisrow{err}] {ALEXCIFAR10SNIP.dat};
\addplot [fill=green!10] fill between[of=upper and lower];

\addplot [name path=upper,draw=none] table[x=x,y expr=\thisrow{y}+\thisrow{err}] {ALEXCIFAR10SYNFLOW.dat};
\addplot [name path=lower,draw=none] table[x=x,y expr=\thisrow{y}-\thisrow{err}] {ALEXCIFAR10SYNFLOW.dat};
\addplot [fill=violet!10] fill between[of=upper and lower];

\addplot [name path=upper,draw=none] table[x=x,y expr=\thisrow{y}+\thisrow{err}] {ALEXCIFAR10DGP.dat};
\addplot [name path=lower,draw=none] table[x=x,y expr=\thisrow{y}-\thisrow{err}] {ALEXCIFAR10DGP.dat};
\addplot [fill=red!10] fill between[of=upper and lower];

\end{axis}
\end{tikzpicture}

\caption{Accuracy vs. sparsity for AlexNet on CIFAR-10. The horizontal line is the accuracy of an unpruned trained network.}
\label{fig:alexnet_cifar10}
\end{figure}

To showcase performance across a wide range of sparsities, we initially evaluated the accuracy of post-trained sub-networks produced by \DRIVE{} and other pruning methods on AlexNet for CIFAR-10, spanning sparsities $\{20\%,30\%,...,98\%\}$. Figure~\ref{fig:alexnet_cifar10} compares the performance of \DRIVE{} and other pruning techniques using AlexNet trained on CIFAR-10 over these sparsity levels\footnote{Due to limited space, we show the figure for one network for a range of sparsities. Our focus is on high sparsity.}. Given the emphasis of our work is in the high sparsity regime, we proceeded to assess the accuracy of post-trained sub-networks crafted by \DRIVE{} and other pruning techniques on VGG-16 and ResNet-18 for CIFAR-10/100 and Tiny ImageNet across competitive sparsities spanning $\{90\%,95\%,…,99.3\%\}$. Furthermore, we measure the wall time needed to derive these sub-networks. A comprehensive set of results is presented in Table~\ref{table:1}.

On the CIFAR-10 dataset, \DRIVE{} consistently outperforms SNIP and SynFlow, especially in sparsity levels exceeding 98\%. \DRIVE{}'s test accuracy demonstrates greater stability (i.e., lower volatility in test accuracy with increasing sparsity), a trait that is closely matched by IMP, an exhaustive pruning method. In some cases, \DRIVE{} even surpasses the accuracy of the IMP. For low sparsity levels (with $\kappa \leq$ 70\%), the performance difference among all methods is minimal, closely aligning with the unpruned baseline. In the mid-sparsity range ($70\% \geq \kappa \geq 90\%$), \DRIVE{} maintains competitive performance, with SNIP achieving similar results. It is noteworthy that SNIP experiences layer collapse in VGG-16 on CIFAR-10 when $\kappa \geq$ 99\%. For AlexNet on CIFAR-10, SNIP outperforms the unpruned baseline until $\kappa \leq$ 80\%. Both IMP and SynFlow outperform the unpruned baseline until $\kappa \leq$ 94\%. \DRIVE{}, however, outperforms the unpruned baseline until $\kappa \leq$ 98\%.

For ResNet-18 on CIFAR-100, \DRIVE{} consistently exhibits superior performance over both SNIP and SynFlow across varying sparsity levels. Until $\kappa \leq 97\%$, the results of SNIP closely align with \DRIVE{}. However, beyond this threshold, the accuracy of SNIP undergoes a substantial reduction, especially when $\kappa > 98\%$. In the regime of $\kappa \leq 90\%$, \DRIVE{} outperforms IMP. However, for sparsity levels exceeding this, IMP marginally outperforms \DRIVE{}. For VGG-16 on CIFAR-100, distinctions among the methods are relatively minor up to $\kappa \leq 80\%$, with all techniques achieving comparable accuracy to the unpruned network. Significantly, \DRIVE{} consistently outperforms SynFlow across all sparsity levels. While SynFlow's performance remains competitive, especially up to $\kappa \leq 98\%$, SNIP experiences a catastrophic drop at $\kappa \geq 99\%$, with its accuracy crashing to just 1\%, which is a clear indication of layer collapse. In contrast, \DRIVE{} sustains its accuracy better, even at elevated sparsity levels, outlasting both SNIP and SynFlow.

For ResNet-18 on Tiny ImageNet, in the low sparsity regime ($\kappa \leq$ 70\%), the  performance difference among all methods is minimal. \DRIVE{}
consistently outperforms SNIP and SynFlow across $\kappa \geq$  90\%. \DRIVE{} offers comparable accuracy to IMP for 90 $ \geq \kappa \geq$  98\%, after which IMP takes the lead. For VGG-16 on Tiny ImageNet, until $\kappa \leq$ 80\%, there is a minimal performance difference among all methods. \DRIVE{}
consistently outperforms SNIP and SynFlow across $\kappa \geq$  90\%. While IMP marginally outperforms \DRIVE{} from 90\%$ \geq \kappa >$ 98\%, after which \DRIVE{} outperforms IMP for $\kappa \geq$  99\%. 

\definecolor{colorIMP}{RGB}{57,106,177}
\definecolor{colorSNIP}{RGB}{62,150,81}
\definecolor{colorSynFlow}{RGB}{128, 0, 128}
\definecolor{colorDRIVE}{RGB}{204,37,41}
\begin{figure}[t]
\centering
\begin{tikzpicture}
    \begin{groupplot}[
        group style={
            group size=1 by 2, 
            xlabels at=edge bottom,
            ylabels at=edge left,
            vertical sep=2 cm,
        },
        ybar,
        width = 8.5 cm,
        height=0.25\textwidth, 
        symbolic x coords={CIFAR-10, CIFAR-100, Tiny ImgNet, ImgNet},
        xtick=data,
        ylabel={Total time (hrs)},
        enlarge x limits=0.24,
        ymin=0,
        legend style={at={(0,1)}, anchor=north west},
        x tick label style={font=\footnotesize, rotate=20},
        nodes near coords,
        every node near coord/.append style={font=\scriptsize},
        nodes near coords align={vertical},
        point meta=explicit symbolic 
    ]

    \nextgroupplot[title={VGG-16}, ymax=25]
    \addplot[color=colorIMP, fill=colorIMP, bar width=0.2cm] coordinates {(CIFAR-10, 3.833) [4] (CIFAR-100, 4.2) [4] (Tiny ImgNet, 19.25) [19]};
    \addplot[color=colorSNIP, fill=colorSNIP, bar width=0.2cm] coordinates {(CIFAR-10, 0.5006) [1] (CIFAR-100, 0.7) [1] (Tiny ImgNet, 1.75) [2]};
    \addplot[color=colorSynFlow, fill=colorSynFlow, bar width=0.2cm] coordinates {(CIFAR-10, 0.5008) [1] (CIFAR-100, 0.7) [1] (Tiny ImgNet, 1.751) [2]};
    \addplot[color=colorDRIVE, fill=colorDRIVE, bar width=0.2cm] coordinates {(CIFAR-10, 0.5556) [1] (CIFAR-100, 0.764) [1] (Tiny ImgNet, 1.777) [2]};

    \nextgroupplot[title={ResNet-18}, ymax=150]
    \addplot[color=colorIMP, fill=colorIMP, bar width=0.2cm] coordinates {(CIFAR-10, 7) [7] (CIFAR-100, 7.333) [7] (Tiny ImgNet, 36.36) [36] (ImgNet, 130.93) [131]};
    \addplot[color=colorSNIP, fill=colorSNIP, bar width=0.2cm] coordinates {(CIFAR-10, 1.167) [1] (CIFAR-100, 1.222) [1] (Tiny ImgNet, 3.306) [3] (ImgNet, 11.9) [12]};
    \addplot[color=colorSynFlow, fill=colorSynFlow, bar width=0.2cm] coordinates {(CIFAR-10, 1.1675) [1] (CIFAR-100, 1.223) [1] (Tiny ImgNet, 3.306) [3] (ImgNet, 11.9) [12]};
    \addplot[color=colorDRIVE, fill=colorDRIVE, bar width=0.2cm] coordinates {(CIFAR-10, 1.2125) [1] (CIFAR-100, 1.28) [1] (Tiny ImgNet, 3.325) [3] (ImgNet, 12.03) [12]};
    
    \end{groupplot}
\end{tikzpicture}
\medskip 
\textcolor{colorIMP}{\rule{2mm}{2mm}} IMP \hspace{1em}
\textcolor{colorSNIP}{\rule{2mm}{2mm}} SNIP \hspace{1em}
\textcolor{colorSynFlow}{\rule{2mm}{2mm}} SynFlow \hspace{1em}
\textcolor{colorDRIVE}{\rule{2mm}{2mm}} DRIVE

\caption{Total pruning and training time of different methods.}
\label{fig:total_time_pruning_training_comparison}
\end{figure}
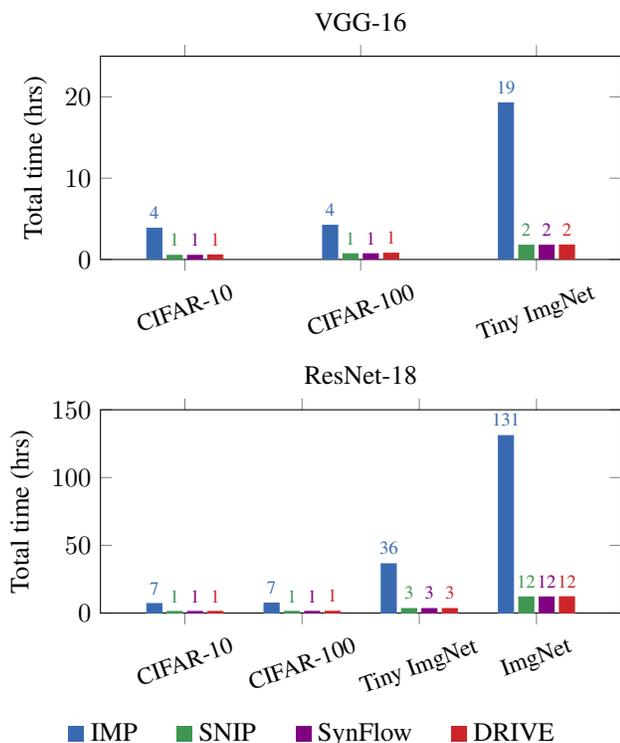

For ResNet-18 on ImageNet, Table~\ref{table:imagenet_resnet18} shows that at lower sparsity levels (up to 70\%), the performance difference between IMP, \DRIVE{}, SNIP, and SynFlow is marginal. However, \DRIVE{} consistently outperforms both SNIP and SynFlow in terms of accuracy for higher sparsities of 90\% and 95\%. While the unpruned model maintains a stable accuracy of 54.11\%, the accuracy of \DRIVE{} decreases less steeply from 49.33\% at 70\% sparsity to 33.16\% at 95\% sparsity. In comparison, IMP shows a similar trend for accuracy reduction, closely following \DRIVE{} at lower sparsities. SNIP and SynFlow have a larger drop in performance as sparsity increases.

\DRIVE{} delivers sparse trainable networks with high accuracy but also expedites pruning when compared to IMP by being 43$\times$ to 869$\times$ faster. We also consider the total time, including both pruning and training in Figure~\ref{fig:total_time_pruning_training_comparison}. \DRIVE{} achieves a runtime comparable to rapid pruning techniques, such as SNIP and SynFlow. In short, \DRIVE{} bridges the gap between the speed of pruning and the accuracy of the sparse networks produced by exhaustive pruning.

\begin{table} [h]
\caption{Test accuracies and pruning time of ResNet-18 on ImageNet for different sparsity levels. Bold values indicate the highest accuracy for the respective sparsity level and model among the early pruning techniques: SNIP, SynFlow, and \DRIVE{}.}
\begin{center}
\scalebox{0.8}{
\begin{tabular}{lccccc}
\hline
Sparsity & 70\% & 80\% & 90\% & 95\% & Pruning Time (s) \\ \hline
Unpruned & \multicolumn{4}{c}{54.11}  \\
IMP & 48.30 & 45.58 & 40.21 & 33.94 & 428.5\(\times 10^3\) \\
SNIP & 48.90 & 45.51 & 36.62 & 21.83 & 3 \\
SynFlow & 48.39 & 45.74 & 39.38 & 31.27 & 4 \\
DRIVE & \textbf{49.33} & \textbf{46.49} & \textbf{40.97} & \textbf{33.16} & 493 \\ \hline
\end{tabular}}
\end{center}
\label{table:imagenet_resnet18}
\end{table}

\section{Conclusions}
\label{sec:discussion}
This paper highlights a trade-off between accuracy and pruning time in network pruning techniques. Iterative Magnitude Pruning (IMP) achieves high accuracy in sparse networks but is computationally intensive. In contrast, methods like SNIP and SynFlow, while faster, fall short in accuracy at high sparsity levels.

To address this challenge, we introduce Dual Gradient-Based Rapid Iterative Pruning (\DRIVE{}), employing a novel dual gradient criterion that combines \textit{connection sensitivity} and \textit{convergence sensitivity}. This new pruning technique not only efficiently identifies and removes redundant parameters but also preserves parameters not yet converged that may have potential future relevance. Consequently, \DRIVE{} attains desired pruning efficiency with substantially lower computational demands compared to the exhaustive train-prune-reset cycles of IMP.

Empirical tests show \DRIVE{} surpasses SNIP and SynFlow in accuracy across various networks and datasets, bridging the gap between exhaustive and rapid pruning methods. Though IMP occasionally produces marginally better results, \DRIVE{} efficiently generates high-quality, sparse trainable networks, achieving speeds that are at least 43$\times$ and up to 869$\times$ faster than IMP. This underscores the benefits of \DRIVE{} in rapidly developing high-calibre sparse networks and offering a viable solution to addressing the energy challenge in training large-scale models by leveraging sparsity from the onset.

\nocite{langley00}

\bibliography{paper-v1}
\bibliographystyle{icml2023}

\end{document}